\documentclass[conference]{IEEEtran}
\IEEEoverridecommandlockouts

\usepackage{cite}
\usepackage{amsmath,amssymb,amsfonts}
\usepackage{algorithmic}
\usepackage{graphicx}
\usepackage{textcomp}
\usepackage{xcolor}
\usepackage{array}
\usepackage{multirow}
\usepackage{cleveref}
\usepackage{url}

\def\baselinestretch{0.975}

\newif\ifanonymous
\anonymousfalse

\def\BibTeX{{\rm B\kern-.05em{\sc i\kern-.025em b}\kern-.08em
    T\kern-.1667em\lower.7ex\hbox{E}\kern-.125emX}}

\begin{document}

\title{Revisiting the Relation Between Language Model Perplexity and ASR Word Error Rate for Modern End-to-End Speech Recognition}

\ifanonymous
\author{\IEEEauthorblockN{Anonymous Author(s)}}
\else
\author{
\IEEEauthorblockN{
Mohammad Zeineldeen\IEEEauthorrefmark{1}\IEEEauthorrefmark{2},
Albert Zeyer\IEEEauthorrefmark{1}\IEEEauthorrefmark{2},
Haoran Zhang\IEEEauthorrefmark{2},
Robin Schmitt\IEEEauthorrefmark{1}\IEEEauthorrefmark{2},
Ralf Schlüter\IEEEauthorrefmark{1}\IEEEauthorrefmark{2},
Hermann Ney\IEEEauthorrefmark{1}\IEEEauthorrefmark{2}
}
\IEEEauthorblockA{\IEEEauthorrefmark{1}
\textit{AppTek.ai GmbH, Aachen, Germany}}
\IEEEauthorblockA{\IEEEauthorrefmark{2}
\textit{Machine Learning and Human Language Technology Group,
Faculty of Computer Science,}\\
RWTH Aachen University
Aachen, Germany\\
\{zeineldeen, zeyer, schmitt, schlueter, ney\}@ml.rwth-aachen.de}
}
\fi

\maketitle

\begin{abstract}
Language model (LM) perplexity (PPL) has historically been used as a proxy for
automatic speech recognition (ASR) word error rate (WER),
with prior work reporting an approximately linear relation in log-log space.
Modern end-to-end ASR systems challenge this assumption because they already
contain internal language modeling capacity,
are often evaluated without external language models, and
can now be combined with neural LMs and large language models (LLMs) through
different recognition strategies. This paper revisits the relation between
PPL and WER for modern ASR systems. We study whether external LMs still
improve current end-to-end ASR systems, whether the PPL-WER relation
remains linear in log-log space, how encoder context length
affects this relation, and how LLM perplexities fit into the trend observed
for standard neural LMs. We further investigate internal language modeling (ILM)
in attention-based encoder-decoder systems and show that ILM subtraction changes
the observed PPL-WER relation, indicating that the decoder's internal LM must
be considered when interpreting the effect of external LM quality.
\end{abstract}


\section{Introduction \& Related Work}

Language models (LMs) have long been central to automatic speech recognition
(ASR). Perplexity (PPL) is commonly used as an intrinsic LM metric, tracing
back to its introduction as a measure of speech task difficulty
\cite{jelinek1977:perplexity}. However, ASR performance is measured by word
error rate (WER), which also depends on the acoustic model, decoding strategy,
pruning, and scale tuning. Thus, lower LM perplexity does not automatically
translate into lower WER.
Earlier studies, mostly using n-gram LMs \cite{ney1995:improved_backoff},
reported experimental evidence for a correlation between perplexity and WER
across several ASR tasks
\cite{chen1998:evaluation_metrics_lm,klakow2002:ppl_wer,halpern2016:contextual_prediction_models}.
The work in \cite{klakow2002:ppl_wer} systematically studied this question and found that
WER and PPL are approximately linearly related in log-log space.
The same log-log linear relation was later observed in \cite{irie2020:phd_thesis}
for hybrid ASR models on Quaero and LibriSpeech dev-clean
using both n-gram and long short-term memory (LSTM) LMs \cite{sundermeyer2015:lstm_lm}.
This observation has been influential because it suggests that
relative PPL improvements can be translated into
expected WER improvements through a power-law relation.

Large-scale ASR studies also support stronger external LMs.
The study in \cite{chelba2012:large_scale_lm} showed that increasing LM scale and using
large text corpora yields consistent WER reductions. Later work extended LM
integration to neural LMs and end-to-end ASR. Shallow fusion combines an
external LM with the ASR model during beam search
\cite{kannan2017:shallow_fusion}, while cold fusion incorporates a pretrained
LM during training \cite{sriram2017:cold_fusion}.
Several integration strategies were compared in \cite{toshniwal2018:lm_integration}
for attention-based encoder-decoder (AED) ASR and
found shallow fusion to be a strong and simple
first-pass decoding method across multiple conditions.
Neural Transformer LMs \cite{vaswani2017:attention} have also been applied successfully to
hybrid and end-to-end ASR, including lattice rescoring and shallow fusion
\cite{irie2019:transformer_lm}.

Modern end-to-end ASR systems make the PPL-WER relation less direct. In
Connectionist temporal classification (CTC) \cite{graves2006:ctc}, recurrent
neural network transducer (RNN-T) \cite{graves2012:rnnt}, and AED
\cite{chan2016:listen_attend_spell} systems, acoustic and language modeling
components are no longer cleanly separated as in classical hybrid ASR.
AED and transducer decoders, in particular, learn an
internal LM from paired speech-text training data.
Moreover, \cite{yang2025:ilm_ctc}
shows that the CTC encoder also learns an internal LM, which can
affect the benefit of external LMs specifically on cross-domain tasks.
This internal LM can interact with an external LM during recognition
and can lead to domain-prior mismatch.
Internal LM estimation and subtraction have
therefore been proposed to improve external LM integration
\cite{zeineldeen2021:ilm,zhou2022:lm_integration_rnnt,yang2025:ilm_ctc}.
These methods suggest that the observed benefit of an external LM
may depend not only on the external LM's PPL,
but also on the strength of the internal language model
of the ASR system.

Recent ASR work often reports greedy recognition or beam search without an
external LM, since LM gains can be small for strong end-to-end models and
no-LM recognition is simpler and faster. This makes it unclear whether
external neural LMs still provide meaningful gains and whether PPL predicts
them. The question becomes more important with
large language models (LLMs), which can be used for
rescoring, prompting, correction, or one-pass search integration, but raise
issues such as inference cost and vocabulary mismatch.
Recent work \cite{hori2025:delayed_fusion} addresses some of these
issues by applying LLM scores during one-pass recognition with delayed scoring
and re-tokenization, and shows improvements over CTC and attention-based
baselines.
However, such work typically does not analyze how LLM PPL
fits into the classical PPL-WER relation.

This motivates a systematic re-examination of the PPL-WER relation for modern
ASR. In this work, we study the following questions:
\begin{itemize}
\item Do external neural LMs still improve modern end-to-end ASR over a no-LM baseline?
\item Is the relation between LM perplexity and ASR WER still approximately linear in log-log space?
\item For AED systems, how does internal LM subtraction affect the strength and shape of the perplexity-WER relation?
\item How does encoder context length affect both the absolute gain from external LMs and the correlation between perplexity and WER?
\item How do LLM perplexities fit into the PPL-WER relation observed for standard neural LMs, and do LLMs follow the same trend when used as external LMs?
\end{itemize}

We address these questions across different ASR architectures, encoder context
settings, recognition modes, and LM families.

\section{Problem Formulation}

\subsection{Perplexity and WER}
Let $a_1^S$ denote the label output sequence, excluding special
beginning-of-sentence (BOS) and end-of-sentence (EOS) symbols.
For autoregressive LMs, we augment the
sequence with $a_0=\langle\mathrm{bos}\rangle$ and
$a_{S+1}=\langle\mathrm{eos}\rangle$. Perplexity is computed as
\begin{equation*}
\mathrm{PPL} = \exp \left( - \frac{1}{S+1} \sum_{s=1}^{S+1}
\log p(a_s \mid a_0^{s-1}) \right),
\end{equation*}
where the final prediction is the EOS label. The label $a_s$ can
denote words, single characters,
or subword tokens \cite{sennrich2016:subword_units,kudo2018:sentencepiece}.
PPL values are only directly comparable when the scoring label unit,
normalization (e.g., cased vs. uncased), and evaluation text are fixed.
In experiments where all LMs share the same 10k SentencePiece (SPM)
vocabulary \cite{kudo2018:sentencepiece}, we report
subword-level PPL. When comparing LMs with different tokenizations, such as
LLMs with different vocabularies, we instead report word-level PPL to make the
perplexities comparable across models.

The ASR metric of interest is WER. Following prior work, we
study the relation between LM perplexity and WER in log-log space:
\begin{equation}
\log(\mathrm{WER}) = \alpha \cdot \log(\mathrm{PPL}) + \beta.
\label{eq:ppl_wer_relation}
\end{equation}
The slope $\alpha$ measures how sensitive WER is to relative changes in PPL,
while the shift $\beta$ captures the recognition setup-dependent offset.
For example, if the relation holds locally, reducing PPL by a factor $r$
changes WER by approximately a factor $r^\alpha$.

\subsection{Regression Protocol}
For each ASR condition, we fit \Cref{eq:ppl_wer_relation} using the LM
variants evaluated under the same recognition setup. When the scatter plot
shows a clear change in slope, we additionally fit a piecewise relation with
a fixed split point and report the slopes of both regions.

\section{ASR and LM Integration}

\subsection{CTC ASR}
For an input acoustic sequence $x_1^T$ and output label sequence $a_1^S$, CTC models
the posterior probability by summing over all frame-level alignments $y_1^T$ that
collapse to $a_1^S$. Unlike autoregressive LMs and AED decoders, CTC does not
use an EOS label; its frame-level label set instead includes the blank label.
\begin{equation*}
p_{\mathrm{CTC}}(a_1^S \mid x_1^T) =
\sum_{y_1^T \in \mathcal{B}^{-1}(a_1^S)}
\prod_{t=1}^T p(y_t \mid x_1^T).
\end{equation*}
The model is trained by minimizing $-\log p_{\mathrm{CTC}}(a_1^S \mid x_1^T)$.
In recognition, we approximate the CTC sum over alignments by the best
alignment score. Without an external LM, this gives
\begin{equation*}
\hat{a}_1^{\hat{S}} =
\arg\max_{a_1^S}
\bigg\{
\max_{y_1^T \in \mathcal{B}^{-1}(a_1^S)}
\sum_{t=1}^{T} \log p(y_t \mid x_1^T)
\bigg\}
\end{equation*}
With an external LM, we use frame-synchronous decoding with a log-linear score
over frame-level CTC labels and a sequence-level LM score:
\begin{equation*}
\begin{aligned}
\hat{a}_1^{\hat{S}} = \arg\max_{a_1^S}
\Bigg\{
&\max_{y_1^T \in \mathcal{B}^{-1}(a_1^S)}
\sum_{t=1}^{T}
\Big[
\log p(y_t \mid x_1^T) \\
&\quad - \mu \log p_{\mathrm{prior}}(y_t)
\Big]
+ \lambda \log p_{\mathrm{LM}}(a_1^S)
\Bigg\}
\end{aligned}
\end{equation*}
where $\lambda$ is the LM scale and $\mu$ is the frame-wise prior scale.
The LM state is advanced only when
non-blank labels are emitted; blank labels do not receive an LM score. The LM
history is initialized with the BOS token $a_0$.
The frame-wise CTC prior is estimated from posteriors computed over the
training data using the softmax average method:
$p_{\mathrm{prior}}(y)
= \frac{1}{\sum_n T_n} \sum_n \sum_{t=1}^{T_n}
p(y_t = y \mid x_{n,1}^{T_n}),$
where $n$ indexes training utterances.

\subsection{AED ASR}
Attention-based encoder-decoder (AED) models directly factorize the output
posterior as
\begin{equation*}
p_{\mathrm{AED}}(a_1^S \mid x_1^T) =
\prod_{s=1}^{S+1} p(a_s \mid a_0^{s-1}, x_1^T),
\end{equation*}
where $a_0=\langle\mathrm{bos}\rangle$ and
$a_{S+1}=\langle\mathrm{eos}\rangle$. The encoder maps the acoustic sequence
to hidden representations, and the autoregressive decoder predicts each output
token conditioned on previous tokens and the encoder context. The model is
trained by minimizing the negative log-likelihood of the reference sequence,
including the EOS label.

With an external LM, AED recognition uses a log-linear score
\begin{equation*}
\begin{aligned}
\hat{a}_1^{\hat{S}} = \arg\max_{a_1^S} \big[
&\log p_{\mathrm{AED}}(a_1^S \mid x_1^T)
+ \lambda \log p_{\mathrm{LM}}(a_1^S) \\
&- \gamma \log p_{\mathrm{ILM}}(a_1^S)
\big],
\end{aligned}
\end{equation*}
where $p_{\mathrm{ILM}}$ denotes an estimate of the AED decoder's internal
language model and $\gamma$ is the ILM subtraction scale. In the ILM
experiments, we compare decoding with $\gamma=0$ against decoding with tuned
ILM subtraction using the Mini-LSTM ILM estimate \cite{zeineldeen2021:ilm}.
Here, sequence-level LM probabilities are factorized with the same BOS
and EOS convention: $p_{\mathrm{LM}}(a_1^S)=\prod_{s=1}^{S+1}
p_{\mathrm{LM}}(a_s \mid a_0^{s-1})$.


\section{Experimental Setup}
\subsection{Corpora}
We evaluate two ASR tasks that differ in speaking style, domain coverage, and
the amount of external LM text.

\textbf{LibriSpeech.}
For read English speech, we use the LibriSpeech 960h corpus
\cite{panayotov2015:librispeech} and report results on the standard
dev/test clean/other sets, focusing mainly on dev-other and test-other. The
LibriSpeech LM corpus contains roughly 800M running words, about 80 times
more than the 10M words in the acoustic transcriptions.

\textbf{AppTek Spanish.} For spontaneous Spanish speech, we use an in-house
AppTek mixed-bandwidth task with 8 kHz and 16 kHz data. It contains about
12k hours of speech, 120M transcription words, and 700M external LM training
words. We report average WER over multiple in-house test sets covering
open-domain speech and conversational call-center domains. Recognition scales
are tuned on matching development sets. Text is lowercased and punctuation is
removed.

The external-text/transcription-text ratio is much larger for LibriSpeech
than for AppTek Spanish, approximately 80 versus 6. This is relevant for
interpreting PPL-WER slopes because it affects how much additional information
the external LM provides beyond the ASR training transcriptions.

\subsection{ASR Systems}
Our LibriSpeech CTC ASR model uses a 16-layer Conformer encoder
\cite{gulati2020:conformer}
with 1024 dimensions and CTC auxiliary losses on top of
layers 4, 10, and 16 (406M parameters).
For AED models, we use a 12-layer Conformer encoder with an LSTM decoder
with dimension 1024 (130M parameters).
For AppTek Spanish, the CTC ASR model uses a 20-layer Conformer encoder
with dimension 896 (438M parameters).
All models are trained for 100 epochs on LibriSpeech and
8 epochs on AppTek Spanish using the AdamW optimizer
\cite{loshchilov2019:adamw}.
For both LibriSpeech and AppTek Spanish, the ASR systems use 10k
SPM subword label units and
apply on-the-fly speed perturbation \cite{ko2015:speed_perturbation} and
SpecAugment \cite{park2019specaugment}.

\subsection{Language Models}

We vary LM architecture and size to obtain a broad PPL range. Unless stated
otherwise, LMs use the same 10k SPM vocabulary and text normalization as the
corresponding ASR system.

For LibriSpeech, we train Transformer LMs \cite{vaswani2017:attention} on the
LibriSpeech LM corpus, varying layers from 2 to 96 and dimensions from 128 to
1280. The strongest recognition LM has 32 layers, dimension 1024, and about
422M parameters.

For AppTek Spanish, we train Transformer, LSTM, and n-gram SPM LMs. The LSTM
variants use 2 or 4 layers and dimensions 256, 512, 1024, and 2048. The
n-gram LMs use orders $n \in \{2,3,4,5,6\}$ with different pruning thresholds;
larger thresholds remove more low-impact n-grams and usually increase PPL.

\subsection{Recognition and Tuning}
We tune recognition hyperparameters and scales by grid search on the
corresponding development sets. CTC systems use time-synchronous search
\cite{graves2006:ctc},
whereas AED systems use label-synchronous beam search \cite{chan2016:listen_attend_spell}.
Unless stated otherwise, we use beam size 64. We also check search errors to ensure that
they do not mask the effect of the external LM; they are in the range of
0.1--0.3\% absolute and are therefore negligible for the trends studied here.

\section{Experiments}

\subsection{Perplexity-WER Relation}
\label{subsec:ppl_wer_relation}

\begin{figure*}[t]
\centering
\includegraphics[width=\textwidth]{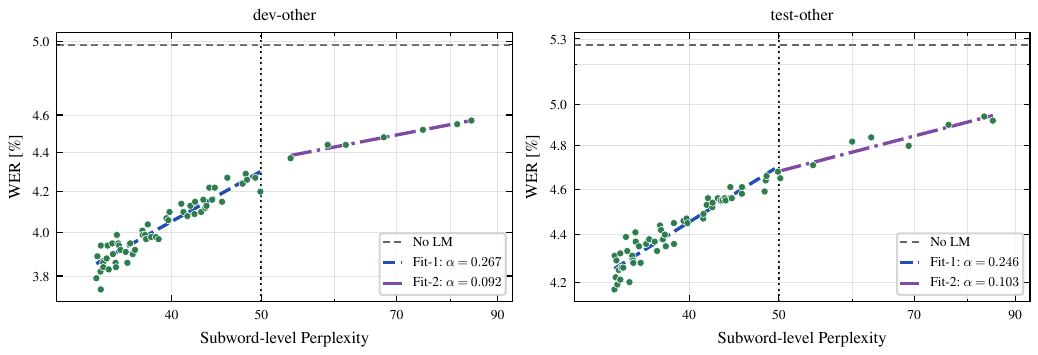}
\caption{
        Relation between subword-level PPL and WER for the LibriSpeech CTC
        system using Transformer LMs.
        The panels show dev-other and test-other.
        Both axes are on the natural log scale.
        We fit $\log(\mathrm{WER}) = \alpha \log(\mathrm{PPL}) + \beta$
        below and above a PPL split point of 50.
        The dashed horizontal line marks CTC recognition without an external
        LM and is not included in the fits.
        On dev-other, Fit-1/Fit-2 have
        $(\alpha,\beta)=(0.267,0.413)/(0.092,1.111)$.
        On test-other, Fit-1/Fit-2 have
        $(\alpha,\beta)=(0.246,0.588)/(0.103,1.142)$.
    }
\label{fig:lbs_ppl_wer}
\end{figure*}

\Cref{fig:lbs_ppl_wer} shows the PPL-WER relation for the LibriSpeech CTC
system with Transformer LMs on the more challenging ``other'' subsets. The PPL-WER trend
is steeper in the low-PPL region and becomes flatter at higher PPLs. With a
split at subword-level PPL 50, the fitted slope on dev-other decreases from
$\alpha=0.267$ to $\alpha=0.092$, while on test-other it decreases from
$\alpha=0.246$ to $\alpha=0.103$. Thus, LibriSpeech shows a clear reduction
in slope, but not a fully saturated high-PPL region. The relation remains
approximately linear up to around this split point, and the high-PPL slopes
remain above zero, consistent with CTC still benefiting from external LM
information beyond this range.

\begin{figure}[t]
\centering
\includegraphics[width=\columnwidth]{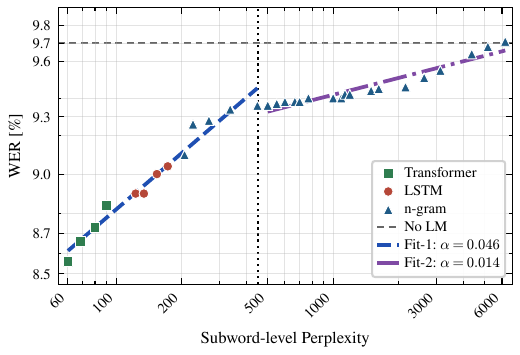}
\caption{
        Relation between subword-level PPL and WER for the AppTek Spanish CTC
        system using Transformer, LSTM, and n-gram LMs.
        Both axes are on the natural log scale.
        We fit $\log(\mathrm{WER}) = \alpha \log(\mathrm{PPL}) + \beta$
        below and above a PPL split point of 450.
        The dashed horizontal line marks recognition without an external LM
        and is not included in the fits.
        Fit-1 has $(\alpha,\beta)=(0.046,1.964)$, and Fit-2 has
        $(\alpha,\beta)=(0.014,2.147)$.
    }
\label{fig:apptek_es_ppl_wer}
\end{figure}

\Cref{fig:apptek_es_ppl_wer} shows the relation between external LM
perplexity and WER for the AppTek Spanish CTC system. The relation is monotonic:
stronger Transformer LMs give the lowest WERs, followed by LSTM and
n-gram LMs. At the same time, the slope becomes smaller in the high-PPL
region. The fit up to a PPL of 450 gives $\alpha=0.046$, whereas the fit
above the split gives $\alpha=0.014$. This suggests a strong flattening at
high perplexities. Compared to LibriSpeech, the AppTek Spanish slopes are
substantially smaller. One likely reason is the ratio between external LM text
and ASR transcription text: for LibriSpeech, the LM corpus is about 80 times
larger than the acoustic transcriptions, whereas for AppTek Spanish this
ratio is about 6. Thus, the external LM can provide more additional
information beyond the transcriptions on LibriSpeech, leading to a stronger
dependence of WER on LM perplexity.

%
%
%

\subsection{Effect of Encoder Context Length}
We next study whether the benefit from an external LM depends on the amount
of encoder context. For this experiment, we use a CTC Conformer model trained
on LibriSpeech 960h and compare greedy CTC decoding, i.e., no external LM,
against recognition with the best Transformer LM available in our
setup. To control the encoder context, we restrict both convolution and
self-attention in the Conformer encoder to a fixed center window. We do not
	use lookahead or history chunks. Thus, each model
has access only to the acoustic frames inside the current center window.

\begin{figure}[t]
\centering
\includegraphics[width=\columnwidth]{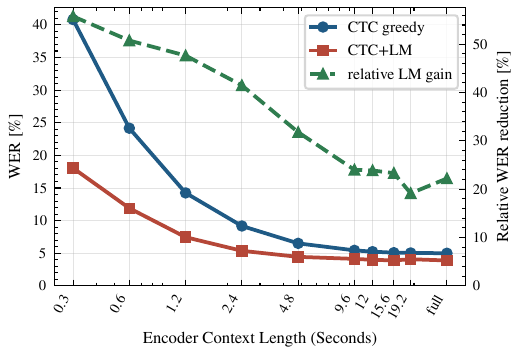}
\caption{
        Effect of encoder context length on CTC recognition with and without
        an external LM. We compare CTC greedy decoding against CTC+LM
        recognition.
        WERs are averaged over LibriSpeech dev-other and test-other.
        The external LM gives larger relative improvements for
        shorter context lengths,
        indicating that LM information becomes more
        important when the encoder has less acoustic context.
    }
\label{fig:encoder_context_lm_gain}
\end{figure}

\Cref{fig:encoder_context_lm_gain} shows the average WER on LibriSpeech
dev-other and test-other for different center-window sizes. The results show
a clear interaction between encoder context and external LM gain. With full
encoder context, the Transformer LM improves WER from 5.00\% to 3.88\%,
corresponding to a relative improvement of 22.2\%. When the center window is
reduced to 0.3 seconds, greedy CTC decoding degrades strongly from 5.00\% to
40.83\%, while CTC+LM recognition reduces WER to 18.05\%. This is a
much larger relative improvement of 55.8\%. Similar behavior is observed for
0.6- and 1.2-second windows, where the LM gives 50.8\% and 47.7\% relative
improvements, respectively.

These results suggest that the external LM becomes more important when the
encoder has limited acoustic context. In the short-context regime, the CTC
model has less information to disambiguate acoustically plausible hypotheses,
and the external LM can compensate for part of this missing context during
search. This may also be interpreted as indirectly weakening the encoder's
internal language modeling capacity \cite{yang2025:ilm_ctc},
which makes the external LM more
beneficial. However, the LM does not fully recover the full-context baseline for
very short windows. For example, the 0.3 second model with an LM still has
18.05\% WER, which is far above the 3.88\% WER of the full-context model with
an LM.

The effect becomes smaller as the center window increases. Around 4.8 seconds
and beyond, the gap between chunked and full-context recognition is much
smaller, and the relative LM gain also decreases. For example, the LM improves
the 4.8 second model by 31.8\% relative, but the gain decreases to about
24\% for 9.6--15.6 second windows, close to the 22.2\% gain observed for the
full-context model. This indicates that once the encoder sees several seconds
of speech, the external LM is still useful but no longer compensates for a
large context deficit.


\subsection{Effect of Temperature-Based LM Probability Smoothing}
\label{subsec:smoothing}

We smooth the external LM distribution by a softmax \emph{temperature} $T$,
i.e., we score the LM as $\log \operatorname{softmax}(z / T)$ from the LM logits $z$,
so that $T > 1$ flattens the distribution and $T < 1$ sharpens it.
We apply this to the LibriSpeech CTC system with a Transformer LM,
sweep $T$, re-tune the log-linear fusion scale for each $T$,
and report the subword-level perplexity and the WER.

\begin{table}[t]
\caption{Effect of LM temperature (T) smoothing for the LibriSpeech CTC system on
LibriSpeech dev-other and test-other. Decoding scales are re-tuned for each
temperature.}
\centering
\small
\begin{tabular}{|c|c|c|c|c|}
\hline
\multirow{2}{*}{$T$} & \multicolumn{2}{c|}{dev-other} & \multicolumn{2}{c|}{test-other} \\
\cline{2-5}
 & PPL & WER[\%] & PPL & WER[\%] \\
\hline
0.50 & 135.1 & 3.81 & 134.9 & 4.14 \\
\hline
0.70 & 51.2  & 3.79 & 51.1  & 4.09 \\
\hline
0.85 & 37.9  & 3.81 & 37.6  & 4.05 \\
\hline \hline
1.00 & 34.1  & 3.75 & 33.8  & 4.02 \\
\hline \hline
1.20 & 35.9 & 3.71 & 35.6  & 4.06 \\
\hline
1.50 & 49.2 & 3.86 & 48.5  & 4.10 \\
\hline
2.00 & 100.2 & 3.87 & 98.7 & 4.05 \\
\hline
\end{tabular}
\label{tab:smoothing}
\end{table}

\Cref{tab:smoothing} shows a controlled case where PPL and WER decouple.
On dev-other, subword-level PPL varies by a factor of 3.96 across the
temperature sweep, from 34.1 to 135.1, while WER only changes between
3.71\% and 3.87\%.
On test-other, subword-level PPL varies by a factor of 3.99, from 33.8 to 134.9, while
WER remains between 4.02\% and 4.14\%. The best PPL also does not always
correspond to the best WER: on dev-other, the lowest PPL is obtained at
$T=1.0$, whereas the lowest WER is obtained at $T=1.2$.

This suggests that PPL changes caused by post-hoc probability smoothing
mainly reflect changes in calibration or entropy, not necessarily changes in
the ranking of recognition hypotheses. After re-tuning the decoding scales,
WER becomes nearly insensitive to these PPL changes. Therefore, PPL-WER
comparisons should distinguish between genuinely different LMs and post-hoc
transformations of the same LM distribution.

\subsection{Effect of Internal LM Subtraction on PPL-WER Relation for AED ASR Models}
\label{subsec:ilm_effect}

\begin{figure*}[t]
\centering
\includegraphics[width=\textwidth]{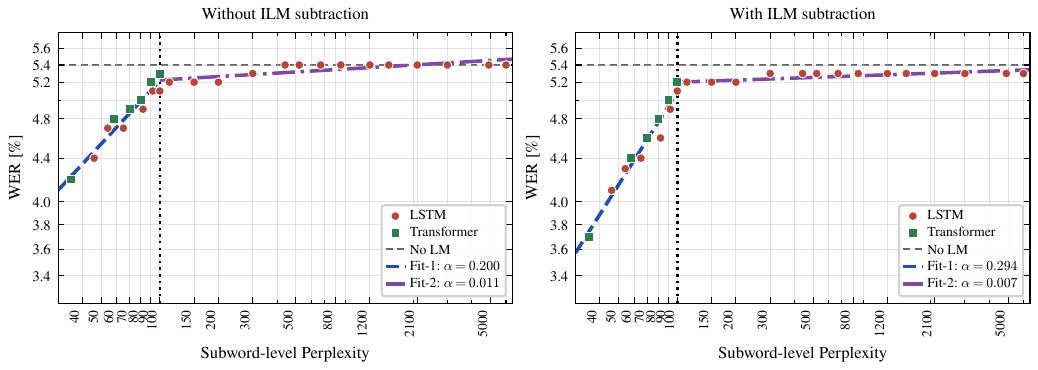}
\caption{
        Relation between subword-level PPL and WER for a Conformer AED model
        on LibriSpeech dev-other, without and with ILM subtraction.
        Both axes are on the natural log scale.
        We fit $\log(\mathrm{WER}) = \alpha \log(\mathrm{PPL}) + \beta$
        below and above a PPL split point of 100.
        The dashed horizontal line marks recognition without an external LM
        and is not included in the fits.
        Without ILM subtraction, Fit-1/Fit-2 have
        $(\alpha,\beta)=(0.200,0.732)/(0.011,1.603)$.
        With ILM subtraction, Fit-1/Fit-2 have
        $(\alpha,\beta)=(0.294,0.275)/(0.007,1.618)$.
    }
\label{fig:att_wer_ppl_ilm_comparison}
\end{figure*}

For attention-based encoder-decoder (AED) models, the decoder learns an
internal language model (ILM) \cite{zeineldeen2021:ilm}.
The ILM acts as a strong prior that can reduce the gain from external LMs
and also weaken the correlation between perplexity and WER.
Thus, we want to investigate how the perplexity-WER relation changes
when we apply ILM subtraction during recognition.
We use the Mini-LSTM ILM \cite{zeineldeen2021:ilm} as the ILM estimation method.
\Cref{fig:att_wer_ppl_ilm_comparison} shows
the correlation between subword-level perplexity and WER using
a Conformer AED model on the LibriSpeech dev-other set without and
with ILM subtraction, respectively.
Both axes are logarithmic, and the regression is performed using natural
logarithms.
In both cases, the relation is described by two regimes.
For perplexity below approximately 100,
the relation is approximately linear in log-log space.
However, with higher perplexities,
the slope is close to 0, indicating a saturation effect
and a weak correlation between perplexity and WER.
This contrasts with the LibriSpeech CTC experiment, where the high-PPL slope
is reduced but remains non-negligible over the evaluated range.

Applying ILM subtraction increases the slope from
0.19 to 0.29 in the low-perplexity regime.
This indicates that ILM subtraction strengthens the
correlation between perplexity and WER,
and also increases the gain from external LMs,
so it is an important factor when studying AED systems.

Moreover, we computed the ILM perplexity of the model
on the dev-other set, and it is 110,
which is close to the empirical transition point
between the low-perplexity and high-perplexity regimes.
This suggests that the saturation effect is likely due to the ILM
dominating the predictions when the external LM perplexity
is higher than the ILM perplexity, which makes the
fitted slope close to 0.

\subsection{How does LLM PPL fit into the PPL-WER relation?}

\begin{figure}[t]
\centering
\includegraphics[width=\columnwidth]{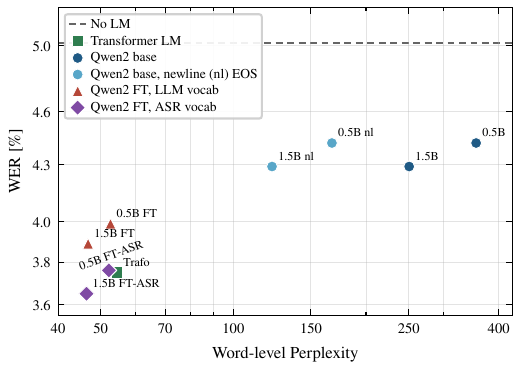}
\caption{
        Relation between word-level PPL and WER for LLM-based recognition on
        the LibriSpeech CTC system, evaluated on dev-other.
        In the legend, ASR vocab denotes the 10k SentencePiece vocabulary of
        the ASR system.
        Both axes are on the natural log scale.
        The dashed horizontal line marks recognition without an external LM.
    }
\label{fig:llm_ppl_wer}
\end{figure}

We next compare standard neural LMs with pretrained text-based Qwen2 LLMs \cite{yang2024:qwen2}
on the LibriSpeech CTC system. Because the LLMs and ASR system may use
different tokenizations, we report word-level PPL in this experiment. We
evaluate Qwen2 base models, variants where newline is used as the EOS label, and
fully fine-tuned models on LibriSpeech transcriptions and external LM text.
For fine-tuning with the ASR SPM vocabulary, we initialize SPM embeddings
from the LLM vocabulary following \cite{deng2025:transducer_llama}: matching
labels reuse the corresponding embedding, while non-matching labels are
initialized by averaging the embeddings of their LLM-tokenized pieces. This
full fine-tuning setup follows our previous work \cite{schmitt2026:llms_speech}.
For recognition, models with the ASR SPM vocabulary use time-synchronous
one-pass search. Models with the original LLM vocabulary use delayed fusion
\cite{hori2025:delayed_fusion}, where LLM scores are fused every 10 frames.
We tune the beam size and observe saturation at beam size 64.

\Cref{fig:llm_ppl_wer} shows that word-level PPL for LLMs must be interpreted
carefully. Word-level PPL is on a different scale from the subword-level PPL
used in the previous experiments, since the LibriSpeech word vocabulary has
about 200k entries and is much larger than the 10k SPM vocabulary.
Thus, the absolute PPL values in this figure should not be compared directly
to the subword-level PPL-WER curves. Under the common word-level scoring used
here, the base Qwen2 models have higher PPL than the Transformer LM,
e.g. 250--356 versus 54.5, but still improve over no-LM recognition, reducing
WER from 5.02\% to 4.29--4.42\%.

The EOS choice mainly affects PPL, not WER. Using newline as EOS reduces PPL
substantially, e.g. from 355.5 to 167.4 for Qwen2-0.5B and from 250.6 to
122.4 for Qwen2-1.5B, while WER remains unchanged at 4.42\% and 4.29\%,
respectively. This suggests that EOS handling can strongly affect intrinsic
PPL without changing the recognition ranking.
Across comparable settings, the larger 1.5B model consistently gives lower
WER than the 0.5B model.

Fine-tuning moves the LLMs closer to the standard LM region under word-level
scoring, but vocabulary and integration still matter. With the original LLM
vocabulary, fine-tuned Qwen2 models reach PPLs comparable to the Transformer LM
but remain worse in WER. In contrast, fine-tuning with the ASR SPM vocabulary
gives the best LLM results: Qwen2-0.5B-FT-SPM matches the Transformer LM
(3.76\% vs. 3.75\%), and Qwen2-1.5B-FT-SPM improves further to 3.65\%.
Thus, this experiment is best interpreted as showing the effect of
tokenization, fine-tuning, and search integration on how LLM PPL relates to
WER, rather than as a direct comparison to the subword-level PPL-WER trends.

\section{Conclusion}
We revisited the relation between language model (LM) perplexity (PPL) and
automatic speech recognition (ASR) word error rate (WER) for modern
end-to-end ASR. The relation remains informative, but is not always a single
global log-log trend: several setups show a low-PPL region with larger WER
sensitivity and a high-PPL region with smaller slope. For Connectionist
temporal classification (CTC), external LMs still help, especially with
restricted encoder context. For attention-based encoder-decoder (AED) ASR, the
internal language model (ILM) affects how external LM quality maps to WER, and
ILM subtraction increases the low-PPL slope. Finally, the large language model
(LLM) experiments show that word-level PPL is harder to interpret across
tokenizations and integration methods: under word-level scoring, Qwen2 LLMs
improve over recognition without an external LM even when its PPL is higher than the
standard LM, newline-based end-of-sentence (EOS) handling changes PPL without
changing WER, and Qwen2 LLMs with the ASR SentencePiece (SPM) vocabulary
give the best LLM results.

\section{Acknowledgments}
The authors gratefully acknowledge the computing time
provided to them at the NHR Center NHR4CES at
RWTH Aachen University (project number p0023999).
This is funded by the Federal Ministry of Education
and Research, and the state governments
participating on the basis of the resolutions
of the GWK for national high performance
computing at universities (\url{www.nhr-verein.de/unsere-partner}).

\section{Generative AI Use Disclosure}
We use LLMs to improve the formulations and grammar of the paper.

\bibliographystyle{IEEEtran}
\bibliography{refs}

\end{document}